\documentclass[letterpaper, 10 pt, conference]{ieeeconf}  % Comment this line out if you need a4paper

\IEEEoverridecommandlockouts                              % This command is only needed if 
\usepackage{cite}
\usepackage{amsmath,amssymb,amsfonts}
\usepackage[dvipsnames]{xcolor}
\usepackage{acro}
\usepackage{hyperref}
\usepackage{booktabs}
\usepackage{subcaption}
\usepackage{pgf}
\usepackage{makecell}

\usepackage{amsthm}
\usepackage[binary-units]{siunitx}[=v2]
\usepackage{mathtools}
\usepackage{pgfplots}
\pgfplotsset{compat=1.8}
\usepgfplotslibrary{statistics}
\usepackage{tikz}
\usepackage{pdfpages}
\usepackage[T1]{fontenc}

\DeclareAcronym{PR}{
    short = TSPR,
    long = Time Series Pattern Recognition,
}
\DeclareAcronym{DR}{
    short = DR,
    long = dimension reduction,
}

\DeclareAcronym{dvb}{
    short = DB,
    long = Dynamic Behavior ,
}
\DeclareRobustCommand\mytikzDotBlue{\tikz[baseline=0.25ex] \fill[blue] (1ex,1ex) circle (0.5ex);}

\DeclareRobustCommand\mytikzDotOrange{\tikz[baseline=0.25ex] \fill[orange] (1ex,1ex) circle (0.5ex);}

\DeclareRobustCommand\mytikzLineOrange{\tikz[baseline=-0.75ex]\draw[line width=2pt,color=orange] (0,0) -- (0.2,0);}
\DeclareRobustCommand\mytikzLineBlue{\tikz[baseline=-0.75ex]\draw[line width=2pt,color=blue] (0,0) -- (0.2,0);}

\definecolor{BlueRawignal}{HTML}{0000a2}
\DeclareRobustCommand\mytikzLineBlueSignalRaw{\tikz[baseline=-0.75ex]\draw[line width=2pt,color=BlueRawignal] (0,0) -- (0.2,0);}

\definecolor{RedRawignal}{HTML}{F28522}
\DeclareRobustCommand\mytikzLineRedSignalDiscretized{\tikz[baseline=-0.75ex]\draw[line width=2pt,color=RedRawignal] (0,0) -- (0.2,0);}

\definecolor{MagentaReducedSignal}{HTML}{800074}
\DeclareRobustCommand\mytikzLineMagentaSignalReduced{\tikz[baseline=-0.75ex]\draw[line width=2pt,color=MagentaReducedSignal] (0,0) -- (0.2,0);}

\definecolor{GreenReducedSignal}{HTML}{06592A}
\DeclareRobustCommand\mytikzLineGreenSignalReduced{\tikz[baseline=-0.75ex]\draw[line width=2pt,color=GreenReducedSignal] (0,0) -- (0.2,0);}

\def\BibTeX{{\rm B\kern-.05em{\sc i\kern-.025em b}\kern-.08em
    T\kern-.1667em\lower.7ex\hbox{E}\kern-.125emX}}

\title{\LARGE \bf
Behavior Forests:\\ Real-Time Discovery of Dynamic Behavior for Data Selection}

\author{Philipp Reis$^{1}$, Philipp Rigoll$^{1}$ and Eric Sax$^{1}$% <-this % stops a space
\thanks{$^{1}$Philipp Reis, Philipp Rigoll and Eric Sax are with the FZI Research Center for Information Technology, 76137 Karlsruhe, Germany
        {\tt\small \{reis,philipp.rigoll,sax\}@fzi.de}}%
}

\begin{document}
%%
%% User Defined Definitions
%%
\theoremstyle{definition}
\newtheorem{definition}{Definition}%[section]
\renewcommand{\theadalign}{vh}
\newcommand{\probP}{\text{I\kern-0.15em P}}

\newcommand{\mathdefault}[1][]{}

\maketitle
\thispagestyle{empty}
\pagestyle{empty}

%%%%%%%%%%%%%%%%%%%%%%%%%%%%%%%%%%%%%%%%%%%%%%%%%%%%%%%%%%%%%%%%%%%%%%%%%%%%%%%%
\begin{abstract}
Automated Driving Systems (ADS) development  relies on utilizing  real-world vehicle data. The volume of data generated by modern vehicles presents  transmission, storage, and computational challenges.  Focusing on \ac{dvb} offers a promising approach to distinguish relevant from irrelevant information for ADS functionalities, thereby reducing data. Time series pattern recognition is beneficial for this task as it can analyze the temporal context of vehicle driving behavior. However, existing state-of-the-art methods often lack the adaptability to identify variable-length patterns or provide analytical descriptions of discovered patterns.
This contribution proposes a Behavior Forest framework for real-time data selection by constructing a Behavior Graph during vehicle operation, facilitating analytical descriptions without pre-training. 
The method demonstrates its performance using a synthetically generated and electrocardiogram data set. An automotive time series data set is used to evaluate the data reduction capabilities, in which this method discarded $\SI{96.01}{\percent}$ of the incoming data stream, while relevant \ac{dvb} remain included.
\end{abstract}

%\begin{IEEEkeywords}
%Pattern Recognition, Behavior Forest
%\end{IEEEkeywords}

%%%%%%%%%%%%%%%%%%%%%%%%%%%%%%%%%%%%%%%%%%%%%%%%%%%%%%%%%%%%%%%%%%%%%%%%%%%%%%%%
\section{Introduction}
\label{sec:introduction}

Real-world vehicle data is used in all development phases of  Automated Driving Systems (ADS) ~\cite{Bach2017} including risk analysis~\cite{Patel21}, perception, path planning and control~\cite{VANBRUMMELEN2018} or simulation and critical scenario identification~\cite{Zhang2023}. %muetsch2023 
These possibilities for utilizing vehicle data face the challenge of dealing with the influx of data during vehicle operation. %Especially in machine learning application, which  requires raw senor data.
Modern vehicles equipped with cameras, RADAR, and LiDAR can generate up to $\SI{40}{\giga\byte\per\second}$ of data~\cite{dickert2023}. With an average driving time of one hour per day and approximately 60 million vehicles in Germany, this would result in a data volume of $8,64$ zetta bytes ($\SI{8.64e12}{\giga\byte}$) of data per year, which is an order of magnitude more data than the current capacity of the data center of the European Organisation for Nuclear Research (CERN) can hold~\cite{cernCapacity2024}. Furthermore, recording these data would be associated with enormous transmission, storage, and computation costs, making it economically and technically inefficient or impossible~\cite{Harris22,hofmockel2019}. As a result, recording all these data volumes is impractical, necessitating a selective approach to recording. This begs the question: How can the influx of data be efficiently managed, and how can data recording during vehicle operation be triggered to obtain a relevant data set for ADS development?\newline
In ADS, developers are interested in Dynamic Behavior (DB) data described by time series like velocity or steering signals. Additionally, considering that \ac*{dvb} happens at different frequencies, such as lane changes on highways being more common than evasive behavior, evaluating relevance regarding information redundancy is essential. \newline
The potential of real-time data selection techniques is demonstrated in physics research at the CERN institute~\cite{elvira2022future}. In an experiment conducted with the Large Hadron Collider, $\SI{40}{\tera\byte\per\second}$ of data are generated. However, using run-time selection techniques,  $\SI{99,98}{\percent}$ of data can be removed~\cite{Albrecht2019}.
In the automotive domain, Tesla uses a \textit{shadow-mode}, which records data if a mismatch between a driver and the vehicle trajectory prediction occurs~\cite{Harris22}, which is rather an error detector than a holistic approach for data recording. \newline
\begin{figure}[t]
    \centering
        \includegraphics[page=1,width=\linewidth]{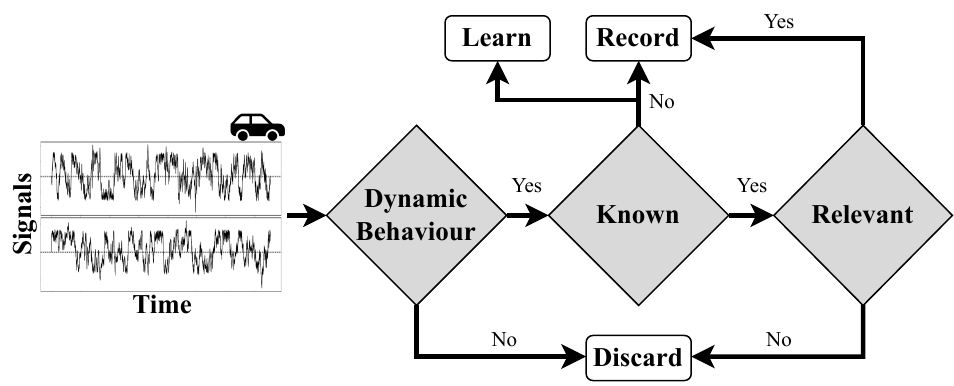}
    \caption{Concept for recording or discarding vehicle data during operation using a  Behavior Forest approach.}
    \label{fig:intro}
\end{figure}
Time Series Analysis provides a variety of methods  to examine sequential data effectively. For instance, it enables the discovery of recurring patterns in time-series data, known as \ac{PR}~\cite{Tanaka2005}, which in this context refers to \ac*{dvb}.
Current approaches to \ac{PR} methods cannot learn what is already known during operation, identify patterns of variable length, or describe the identified patterns analytically.
This contribution addresses these challenges by proposing a Behavior Forest  for \ac*{dvb} discovery of variable length explicitly designed for real-time data selection, see Fig.~\ref{fig:intro}. This entails identifying unseen \ac*{dvb}, learning their representation, and recording them if they are assessed as relevant while disregarding non-dynamic and already known irrelevant dynamic behavior.

\subsection{Related Works}

Methods for \ac{PR}  encompass various approaches, including sliding window techniques, signal thresholding, transformer models, and tree graphs. Especially, methods reliant on signal thresholds for detecting the initiation and termination of \ac*{dvb} are straightforward to implement, such as thresholds applied to lateral and longitudinal acceleration~\cite{Paefgen12} or rotational energy~\cite{Johnson2011}. However, these methods often lack detailed characterization of \ac*{dvb} progression and may oversimplify data recording conditions.
A widely adopted strategy for \ac{PR} involves employing a sliding window across the time series. This method  utilizes raw signal values or derived features from rolling time windows, which are subsequently employed in (un)supervised learning models for classification tasks~\cite{Xie2017, Ries2018, Saleh2017}. Despite its efficacy, the sliding window approach necessitates careful consideration of two critical hyperparameters: the \textit{window size}, which dictates the temporal span of a \ac*{dvb}, and the \textit{window overlap}. Suboptimal settings for these parameters, such as zero \textit{window overlap} leading to information loss or maximal \textit{window overlap} yielding meaningless results, have been discussed in~\cite{Keogh2003}.
Recently, transformer architectures with self-attention mechanisms have emerged as promising tools for time series analysis, finding applications in various domains~\cite{zhang2024,wen2023}. However, despite their success in natural language processing tasks, concerns have been raised regarding their applicability in time series analysis due to inherent features like positional encoding and tokenization~\cite{zeng2022}. % which is still in discussion~\cite{YesTransformer2024}. 
Additionally, formalizing discovered patterns interpretatively to extract actionable knowledge and facilitate continual learning processes is challenging, often complicated by issues such as the catastrophic forgetting phenomenon~\cite{luo2023}.
In  time series analysis, tree graphs have been utilized for tasks such as forecasting~\cite{Houssainy2021} and anomaly detection~\cite{Hofmockel2018} through methods like isolation trees. While effective, these approaches typically require training data and are not commonly explicitly employed for pattern identification.
Moreover, none of the aforementioned approaches can learn from identified patterns and organize them meaningfully  to address the challenge of selecting data by a specific data distribution condition of discovered patterns.

\subsection{Contribution and Outline}

The contribution of this paper unfolds in two main aspects:
Firstly, it introduces an approach for \ac{PR} of variable length without pre-training. These patterns are systematically structured analytically  through a bottom-up construction of a Behavior Forest, effectively compressing observed data. This approach represents a continual learning process.
Secondly, it proposes an approach for real-time data selection  by solely recording relevant \ac*{dvb}.
This method is analyzed for \ac{PR} using a synthetic and an electrocardiogram (ECG) data set. 
Additionally, experimental results from an automotive time series data set showcase its efficacy in handling data and facilitating real-time data recording trigger actions based on discovered \ac*{dvb}.

Section~\ref{sec:preleminaries} presents the necessary notation and definitions essential for \ac{PR} in this paper, while Section~\autoref{sec:concept} introduces the conceptual framework for discovering relevant \ac*{dvb}, outlining all its components. Section~\ref{sec:evaluation} thoroughly evaluates and discusses the method. Finally, Section~\ref{sec:conclusion_and_outlook} summarizes the findings of this contribution and provides insights into potential future research avenues.

\section{Preleminaries}\label{sec:preleminaries}
Before delving into the concepts of the proposed method, the following terms are defined:

\begin{definition}[Time Series $\mathcal{T}$~\cite{morchen2006time}]
A d-dimensional time series w.r.t. a series of time points ${\mathbb{T}}=\{t_{1},...,t_{n}\}$ of length $n\in\mathbb{N}$ is $\mathcal{T}=\{(t_{i},y_{i}){\vert y_{i}}= (y_{i,1},...,y_{i,d})^{T}\in\mathbb{R}^{d}, t_{i}\in\mathbb{T}, i=1,\cdots,n\}$. If $d=1$, $\mathcal{T}$ is called \textit{univariate}, for $d > 1$ it is a multivariate time series.
\end{definition}
\begin{definition}[Subsequence $\mathcal{T}_{i,\mathrm{l}}$~\cite{Noering2022}]
A subsequence $\mathcal{T}_{i,l} \in \mathbb{R}^{n\times d+1}$ of a time series $\mathcal{T}$ is a subset of values from $\mathcal{T}$ of length $l$ starting from index $i$. This subsequence $\mathcal{T}_{i,l}$ contains consecutive tuples $\begin{bmatrix}t_{i} &t_{i+1}&\hdots&t_{i+l-1}\end{bmatrix}$.
\end{definition}

\begin{definition}[Pattern $\mathcal{P}$ based on~\cite{Li2012} ]
A time series pattern $\mathcal{P}$ is a group of m (with $m \geq 2$) similar subsequences in a time series $\mathcal{T}$. A subsequence $\mathcal{T}_{i,\mathrm{l}}$ , that is
included in a pattern $\mathcal{P}$, is called a member M. Formally, $\mathcal{P} = [M_1, M_2, \hdots , M_m]$ with each $M_x$ being a subsequence with a starting index i and a length l. This includes a distance threshold $\tau$ with $ dist(M_x , M_y) \leq \tau, \, \forall x, y \in [1, \hdots,m]$.
\end{definition} 

\begin{definition}[Dynamic Behavior (DB)]\label{def:db}
A Dynamic Behavior (DB) is a pattern $\mathcal{P}$ that initiates from and concludes in a stationary  state.
\end{definition}

The goal of the real-time \ac{dvb} discovery is to find the \ac{dvb} during operation and trigger the recording if a  relevance criteria is met.
\section{Concept}
\label{sec:concept}
The proposed method for discovering \ac{dvb} in a continuous data stream comprises two steps: A preprocessing step based on~\cite{Noering2021}, which is designed for discovering vehicle driving patterns involving discretization, hysteresis filtering, dimension, and numerosity reduction \textendash{} Moreover, the construction of a Behavior Forest, which is the main contribution of this paper.

\begin{figure*}[htp]
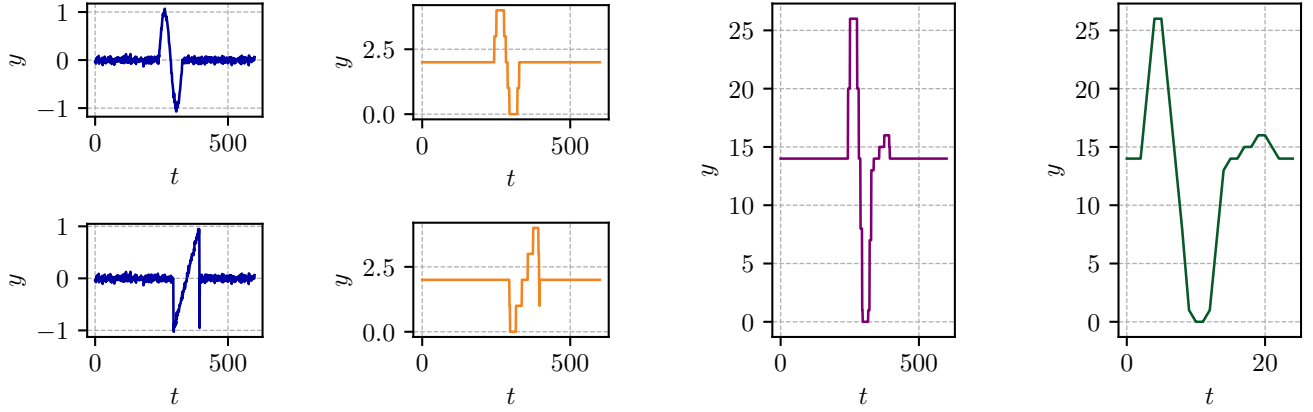

\begin{subfigure}[t]{0.2\textwidth}
    \scalebox{0.97}{\input{graphics/signals_raw.pgf}}
\end{subfigure}
\hspace{0.8cm}
\begin{subfigure}[t]{0.2\textwidth}
    \centering
    \scalebox{0.97}{\input{graphics/signals_discrete.pgf}}
\end{subfigure}
\hspace{0.8cm}
\begin{subfigure}[t]{0.2\textwidth}
    \scalebox{0.97}{\input{graphics/signal_unification.pgf}}
\end{subfigure}
\hspace{0.8cm}
\begin{subfigure}[t]{0.2\textwidth}
    \scalebox{0.97}{\input{graphics/signal_numerosity.pgf}}
\end{subfigure}
\caption{Preprocessing of data. The raw multivariate time signals (left, \mytikzLineBlueSignalRaw) are discretized with predefined breakpoints (middle~left,~\mytikzLineRedSignalDiscretized). Using unification, the multivariate time series is transformed into a univariate time series (middle right,~\mytikzLineMagentaSignalReduced) and finally compressed with a logarithmic numerosity reduction method (right,~\mytikzLineGreenSignalReduced).}
\label{fig:concept}

\end{figure*}

\subsection{Preprocessing}

\subsubsection{\textbf{Discretization}}
Symbolic Aggregate Approximation (SAX) is a method utilized to transform the original signal into a symbolic representation using predefined thresholds, thereby reducing computational resources as demonstrated in~\cite{Lin07}, see Fig.~\ref{fig:concept} (a) to (b). 
\begin{definition}[Discretization]
    Given a natural number $\alpha\geq2$, define $\probP \coloneqq \begin{pmatrix}\frac{1}{\alpha} \vert 1,\hdots ,\alpha \end{pmatrix}$. Applying the quantile function of the standard normal distribution to the probabilities in $\probP$
results in the $\alpha$-quantiles $\beta_j$ with $1\leq j\leq \alpha$ of the standard normal distribution.
The ascending sorted list of these $\alpha$-quantiles  
\begin{equation}\label{eq:breakpoints}
    \mathrm{B} \coloneqq \begin{bmatrix}\beta_0, \beta_1, \hdots ,\beta_{\alpha}  \end{bmatrix}
\end{equation} 
with $\beta_0 = -\infty$ and $\beta_\alpha = +\infty$ are called breakpoints and $\alpha$ is called alphabet size. 
\end{definition}
This way, a time series $\mathcal{T}$ is transformed in a symbolic time series $\mathcal{T}_{\mathrm{s}}$, using the discretization function $f_\mathrm{s}$
\begin{equation}
    f_\mathrm{s}:\mathcal{T}\rightarrow \mathcal{T}_{\mathrm{s}}.
\end{equation}

\subsubsection{\textbf{Hysteresis Filter}}
Signals which are in the value regions of the breakpoints $\beta_i$ from \eqref{eq:breakpoints} are prone to lead to symbolic toggling due to noise. This is eliminated by the hysteresis filter~\cite{Noering2022}.

\subsubsection{\textbf{Dimension Reduction}}
Reducing the dimensionality of multivariate time series increases the computation efficiency of \ac{PR}, which can be rule-based, spectral, probabilistic, and unsupervised learning-based~\cite{petersen24}. Except for rule-based dimension reduction methods, these approaches need data in advance for parameter tuning. Therefore, this paper uses the rule-based method \textit{unification} by mapping all combinations of discretization symbols to a unique symbol using the cartesian product, see Fig.~\ref{fig:concept} (b) to (c). For two discretized signals $B_{\mathrm{1}}$ and $B_{\mathrm{2}}$, this leads to unified discretization breakpoints $B_{\mathrm{u}}$:
\begin{equation}\label{eq:dimReduciton}
    B_{\mathrm{u}} = B_{\mathrm{1}} \times B_{\mathrm{2}}. 
\end{equation}

\subsubsection{\textbf{Numerosity Reduction}}
Numerosity Reduction is a data reduction technique that aims to reduce the number of data points while preserving its essential characteristics, see Fig.~\ref{fig:concept} (c) to (d). The consideration of temporally distorted signals as the same pattern is approached with a logarithmic reduction of the symbol sequence with the base $a$, that is, for each subsequence of the same symbol $\mathcal{S}_{\mathrm{same}}$ leading to a numerosity reduced subsequence $\mathcal{S}_{\mathrm{reduced}}$:
\begin{subequations}\label{eq:numerosity}
\begin{equation}
   L\left(\mathcal{S}_{\mathrm{reduced}} \right) = \lceil \log_a{L(\mathcal{S}_\mathrm{same}} )\rceil
\end{equation}
\end{subequations}
where $L(\mathcal{S}_\mathrm{same})$ is the length of consecutive identical symbols in a subsequence  and  $\lceil X \rceil$ being the ceiling function.

These preprocessing steps reduce the computational complexity of the time series while preserving the pattern characteristics. Based on this preprocessing, the discovery and analytical description take place using the Behavior Forest.

\subsection{Continual \ac{dvb} Discovery with Behavior Forests}

\begin{figure}[!tbp]
  \begin{subfigure}[b]{0.25\textwidth}
    \centering
    \begin{tikzpicture}[scale=0.7] % Adjust the scale as needed
    % Define the series of numbers
    \def\series{{1,1,1,2,3,2,1,1,1,1,1,2,3,4}}
    \def\scaleWidth{0.4} % Define a variable to scale the width between the numbers
    \def\BoxHeight{0.7}
    
    % Draw the series of numbers
    \foreach \x/\y in {0/1, 1/1, 2/1, 3/2, 4/3, 5/2, 6/1, 7/1, 8/1, 9/1, 10/1, 11/2, 12/3, 13/4} {
        \ifnum\x=13
            \draw[fill=ForestGreen] (\x*\scaleWidth,0) rectangle ++(\scaleWidth,\BoxHeight) node[midway] {\y};

        \else
            \draw (\x*\scaleWidth,0) rectangle ++(\scaleWidth,\BoxHeight) node[midway] {\y};
        \fi
    }
    
    % Draw brackets above
    \draw [decorate,decoration={brace,amplitude=5pt},yshift=3pt] (2.25*\scaleWidth,\BoxHeight) -- (6.75*\scaleWidth,\BoxHeight) node [midway,yshift=10pt] {DB};

    \draw [decorate,decoration={brace,amplitude=5pt},yshift=3pt] (10.25*\scaleWidth,\BoxHeight) -- (13.75*\scaleWidth,\BoxHeight) node [midway,yshift=10pt] {DB};
 \node[draw=none] at (0,-1.5) {};  

\end{tikzpicture}
    \caption{Time sequence}
    \label{fig:TimeSequence}
  \end{subfigure}
  \hfill
  \begin{subfigure}[b]{0.2\textwidth}
    \centering
    \def\distance{1}
\def\offset{0.5}
\def\LabelOffset{0.15}
\begin{tikzpicture}[every node/.style={rectangle,rounded corners=0.1cm,draw},level distance=\distance cm]
  \node {1}
    child[->] {node {2} 
      child[->]  {node {3}
          child[->,grow=right] {node[level distance=1.1cm]  {2}
            child[grow=south] {node  {1}}}
          child[->,grow=south, fill=ForestGreen] {node[fill=ForestGreen] {4}}}
    };
 \node[draw=none] at (-\LabelOffset,\offset-\distance) {\footnotesize 2};  
 \node[draw=none] at (-\LabelOffset,\offset-2*\distance) {\footnotesize 2};
 
  \node[draw=none] at (-0.5+\distance,\offset -2*\distance-0.3) {\footnotesize 1};

 \node[draw=none] at (-\LabelOffset,\offset -3*\distance) {\footnotesize 1};  
 \node[draw=none] at (-\LabelOffset +\distance,\offset-3*\distance) {\footnotesize 1};  

\end{tikzpicture}
    \caption{Behavior Tree}
    \label{fig:Behavoir_Tree}
  \end{subfigure}
   \caption{Exemplary construction of a Behaviour Tree  from a time sequence with the indicated \ac{dvb}.}
  \label{fig:BehaviorGraph}
\end{figure}
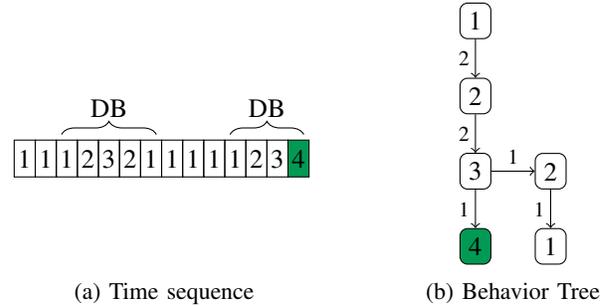

The core of pattern identification lies in the bottom-up construction of a Behavior Tree, illustrated in Fig.~\ref{fig:BehaviorGraph}. The tree structure (right) adapts as the algorithm processes the data stream (left), ultimately contributing to the broader Behavior Forest. 
The algorithm identifies the initiation of a \ac{dvb} when a symbol differs from the two preceding symbols of the same value.
Upon detecting a \ac{dvb} initiation, the construction of a Behavior Tree commences. Subsequently, the first symbol in the \ac{dvb} sequence becomes the root node of this tree.
As subsequent symbols arrive from the data stream, the algorithm traverses the Behavior Tree, searching for a child node of the current node that matches the incoming symbol. If a matching child node exists, the traversal continues down that branch,  indicating that the observed sequence aligns with a previously encountered \ac{dvb}. However, if no matching child node exists, a new child node is created for the incoming symbol, signifying the discovery of a new element within the \ac{dvb} pattern, as indicated in green. The traversal then proceeds down the newly created branch, extending the tree structure.
The traversal continues until a termination condition is met. This can occur when the data stream indicates a steady-state driving scenario identified by three consecutive symbols. Leaf nodes in the Behavior Tree represent the final symbols of a \ac{dvb}.  Besides analytically describing \ac{dvb}, the algorithm tracks the frequency of each \ac{dvb}, indicated by the edge weight of the Behaviour Tree. With each identified \ac{dvb} and its corresponding Behavior Tree construction, the frequency count associated with the leaf node increments. This functionality allows the algorithm to identify \ac{dvb} and quantify their prevalence within the data stream.
Overall, the Behavior Tree construction process enables the algorithm to dynamically capture the structure and occurrence frequency of \ac{dvb} observed in the continuous data stream.

\begin{figure*}[!tbp]
  \begin{subfigure}[b]{0.475\textwidth}
    \centering
    \scalebox{0.95}{\input{graphics/example_motif.pgf}}
    \caption{}
    \label{fig:example_motif}
  \end{subfigure}
  \hfill
  \begin{subfigure}[b]{0.475\textwidth}
    \centering
    \includegraphics[page=1,width=\textwidth]{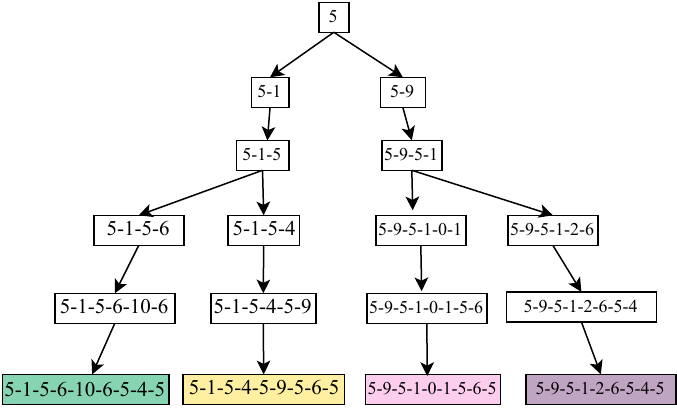}
    \hspace{10mm}
    \caption{}
    \label{fig:treeStructure}
  \end{subfigure}
   \caption{Visualization of the two-dimensional synthetic time series in (a, top), in which signal one is orange \mytikzLineOrange\hspace{0.5mm} and signal two is blue \mytikzLineBlue\hspace{0.5mm}. An extract of the Behavior Forest (b) represents four distinct patterns as a colored leafs. These colored leafs are highlighted according to the pattern in the original time series (a, top), and in the preprocessed symbolic sequence (a, bottom) which is the Behavior Forest algorithm input.}
  \label{fig:motifGraph}
\end{figure*}

\section{Numerical Evaluation}
\label{sec:evaluation}
The evaluation is divided into two sections: the first section assesses the discovery of patterns in two different data sets, while the second part evaluates the application of data reduction in a vehicle speed data set by discovering \ac{dvb} for data selection.
\subsection{Discovery of Patterns}

\subsubsection{Synthetic Time Series}

For a demonstration of \ac{PR} according to Definition \ref{def:db}, a two-dimensional time series with four patterns consisting of alternating sawtooth and sinus wave signals with amplitudes of $a_1=1$ and $a_2=0.6$  with added Gaussian noise are generated, see Fig.~\ref{fig:example_motif} (top). The pattern discovery algorithm uses the parameter according to table \ref{tab:ParamSynth}.
\begin{table}[h] 
    \centering
    \caption{Parameter Values Synthetic Data}
    \begin{tabular}{lr}\label{tab:ParamSynth}
        %\toprule
        \textbf{Parameter} & \textbf{Value} \\
        \midrule
         logarithmic base $a$  & 10 \\
         $B_{\mathrm{1}/\mathrm{2}}$ & $\left[ -0.5, 0.5 \right]$ \\
        \bottomrule
    \end{tabular}
\end{table}
The resulting Behavior Forest contains four leafs indicating four different patterns, as shown in Fig.~ \ref{fig:treeStructure}. The colored patterns  are shown in the original time series and in the preprocessed symbolic sequence. All multivariate patterns are found in the preprocessed data and  represented in the Behavior Graph.

\subsubsection{Electrocardiogram Recording}
\begin{figure}
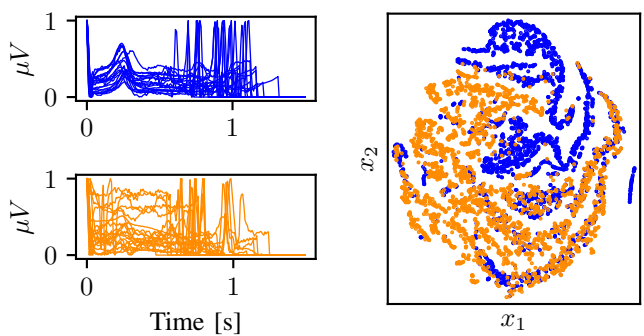

    \begin{subfigure}[b]{0.25\textwidth}
        \centering
        \input{graphics/normal_abnormal_ts.pgf}
    \end{subfigure}
    \begin{subfigure}[b]{0.2\textwidth}
        \centering
        \scalebox{1}{\input{graphics/tsne_noraml_abnormal.pgf}}
  \end{subfigure}
    \caption{Subset of normal \mytikzLineBlue/\mytikzDotBlue \hspace{0.5mm} and abnormal \mytikzLineOrange/\mytikzDotOrange\hspace{0.5mm} ECG time signals (left). Visualizing the features of found patterns (mean, variance, skew, kurtosis, minimum, maximum, median, 25th, and 75th percentiles) in a two-dimensional plane using the t-SNE algorithm (right).}
    \label{fig:tsne_ecg}
\end{figure}

A popular time series for analysis is the open-source electrocardiogram  (ECG) recording \cite{ECGData1992} since the patterns are labeled as normal and abnormal. For evaluation, the behavior Forest method is applied to the normal, see Fig.~\ref{fig:tsne_ecg}  in blue \mytikzLineBlue, and abnormal, see Fig.~\ref{fig:tsne_ecg} in orange \mytikzLineOrange, ECG recording separately. The parameters are set according to table \ref{tab:ParamECG}.
\begin{table}[htbp] 
    \centering
    \caption{Parameter Values ECG Data}
    \begin{tabular}{lr}\label{tab:ParamECG}
        %\toprule
        \textbf{Parameter} & \textbf{Value} \\
        \midrule
         logarithmic base $a$  & 10 \\
         $B$ & $\left[ 0,0.2,0.4,0.6,0.8,1 \right]$ \\
        \bottomrule
    \end{tabular}
\end{table}
The Behavior Forest extracted 80 patterns from normal and 6 patterns from abnormal ECG data set. From each pattern, nine features are extracted which are: mean, variance, skew, kurtosis min, max, median, 25 and 75 percentil. This nine-dimensional feature vector of each pattern is visualized using the t-SNE algorithm into a two-dimensional plane, see Fig.~\ref{fig:tsne_ecg}. A blue dot \mytikzDotBlue \hspace{.5mm} indicates a normal ECG signal whereas an orange dot \mytikzDotOrange \hspace{.5mm} indicates an abnormal ECG signal. It can be seen that some pattern features occur in both signals. However, compared to the normal ECG data set, the abnormal data set has a disturbed ECG signal between $t_\mathrm{start}=0s$ and $t_\mathrm{end}\approx 0.5s$. These resulting patterns lead to the blue dominating pattern features. Consequently, the behavior Forest method discovered patterns that are both present in the normal and abnormal, and patterns that are only present in one of the normal or abnormal ECG data sets, allowing them to be distinguished.

\subsection{Vehicle Data Selection}
To demonstrate the method for real-time data selection, the vehicle speed data set containing 5973 rides with a total length of $9049.3$km  \cite{vehicleSpeedData2023} is used for evaluation. The vehicle speed is generated synthetically using LSTMs with a speed range of $v_{\min}=0$ km/h and $v_{\max}= 130$ km/h. For the discovery of \ac{dvb} with the proposed method, the parameters are set according to table \ref{tab:ParamVehicle}.
\begin{table}[h] 
    \centering
    \caption{Parameter Values Vehicle Data}
    \begin{tabular}{lr}\label{tab:ParamVehicle}
        %\toprule
        \textbf{Parameter} & \textbf{Value} \\
        \midrule
         logarithmic base $a$  & 10 \\
         $B$ & $\left[ 0, 5, 10, 17.5,25, 35, 45, 55, 70 \right]$ \\
        \bottomrule
    \end{tabular}
\end{table}
For evaluation of the data selection, the entire data set is repeated several times to simulate a progression from no prior knowledge to a known driving situation. The evaluation starts with an empty Behavior Forest, representing zero prior knowledge of the course.  The criteria for data recording are:
\begin{itemize}
    \item A leaf is created in the Behavior Forest, which indicates an unseen \ac{dvb}.
    \item A known \ac{dvb} is considered relevant if it has occurred less than five times. 
\end{itemize}
During the initial run, 697 \ac{dvb} were detected, comprising $\SI{8.89}{\percent}$ of the complete data set. However, by the fifth run, all data was discarded, as indicated in Table \ref{tab:saturation}. Consequently, the amount of recorded data decreases progressively, which indicates a learning process of the discovered \ac{dvb} inside the data set. 
\begin{table}[htbp]
    \centering
    \caption{Recording of Vehicle Data}
    \begin{tabular}{cccccc}\label{tab:saturation}
        %\toprule
        \textbf{} & \textbf{Run 1}& \textbf{Run 2}& \textbf{Run 3}& \textbf{Run 4}& \textbf{Run 5}    \\
        \midrule
         Recorded \ac{dvb}   & 697  & 383& 277& 171 & 0  \\
          Recording/Run & $\SI{8.89}{\percent} $ &$\SI{5.02}{\percent} $& $\SI{3.78}{\percent} $& $\SI{2.3}{\percent} $ & $\SI{0}{\percent}$  \\
         Total Recording & $\SI{8.89}{\percent} $ &$\SI{6.96}{\percent} $& $\SI{5.90}{\percent} $& $\SI{4.99}{\percent} $ & $\SI{3.99}{\percent}$  \\
        \bottomrule
    \end{tabular}
\end{table}
Overall, $\SI{3.99}{\percent}$ of the data are recorded with 1528 \ac{dvb} detections. By discarding $\SI{96.01}{\percent}$ of the data stream, the recorded data needs evaluation to determine if they represent the desired \ac{dvb} data. Therefore, the variance of the recorded data is computed and compared against the variance of a sliding window across the data set with $\SI{50}{\percent}$ overlap. The window length is set to the mean of all discovered \ac{dvb}, as depicted in Fig.~\ref{fig:boxplot_variance}. It is evident that the selected recorded data represent time series with high variance, indicating a more dynamic movement. 
\begin{figure}[b]
    \centering
        \resizebox{.95\linewidth}{!}{\begin{tikzpicture}
  \begin{axis}
    [
    xlabel= Signal Variance,
    ytick={1,2},
    yticklabels={Sliding Window, DB},
    y post scale=0.25,
    ]
    \addplot+[
    boxplot prepared={
      median=4.927877,
      upper quartile=14.863230,
      lower quartile=3.062754,
      upper whisker=32.540749,
      lower whisker=0
    },
    color = black,
    ] coordinates {};
    \addplot+[
    boxplot prepared={
      median= 29.809971,
      upper quartile=160.050794,
      lower quartile=6.473433,
      upper whisker=  387.055833,
      lower whisker=0
    },
    color = black,
    ] coordinates {};
  \end{axis}
\end{tikzpicture}}
        \caption{Comparison of the signal variance of the recorded \ac{dvb} data and sliding window across the whole data set.  }
        \label{fig:boxplot_variance}
\end{figure}
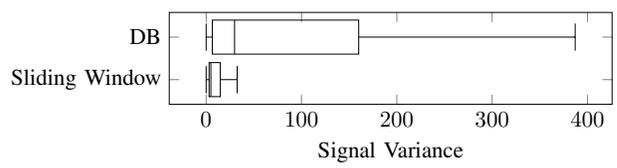
\begin{figure*} 
    \centering
    \includegraphics[width=0.9\textwidth]{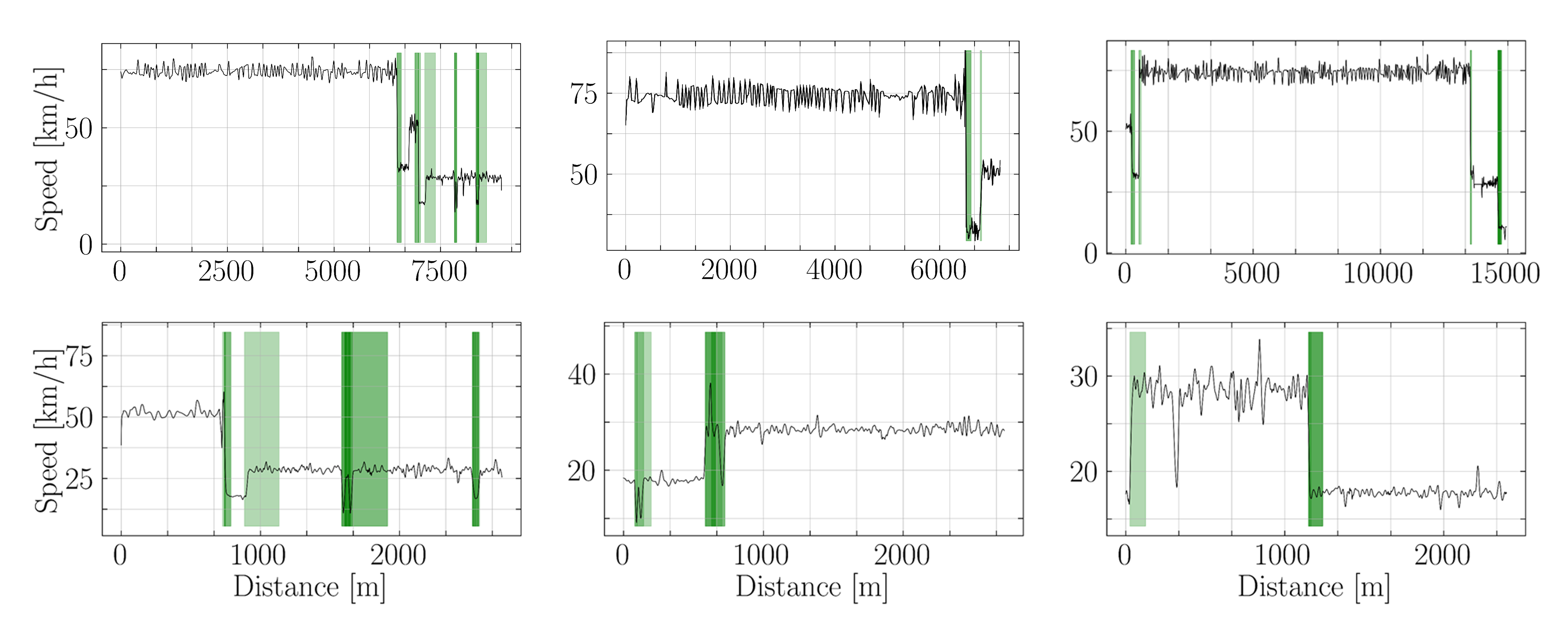}
    \caption{Six different driving speed signals from the data set, with the variable-length \ac{dvb} highlighted in green, recorded using the Behavior Forest method. Darker green highlights indicates a detected sub-pattern within a pattern.}
    \label{fig:ts_motifs}
\end{figure*}
Additionally, six of the 5973 rides are shown in Fig.~\ref{fig:ts_motifs} with the recorded data, in which the detected \ac{dvb} are marked in green. It can be seen that both dynamic and variable length \ac{dvb} were recorded, which underlines the performance of the proposed method.

\subsection{Discussion and Limitations}
Besides the effectiveness of the proposed method, two other aspects need to be discussed.
First, the criteria for relevant data is set to a threshold of five occurrences in the application of data selection. For real-world applications, more advanced criteria should be considered.
Second, the proposed method utilizes two hyperparameters: discretization breakpoints and logarithmic reduction basis, which must be set.  Expert knowledge is necessary to set these hyperparameters properly. However, in applications like \ac{dvb} detection, where ranges of the dynamical system are known, determining suitable breakpoints $\beta_i$ and temporal scaling becomes feasible. Therefore, the limitation of this method lies only in applications where no prior information about the data is available.

\section{Conclusion \& Outlook}
\label{sec:conclusion_and_outlook}
Compared to state-of-the-art methods for discovering Dynamic Behavior (DB) in  time series data, the proposed approach excels in several aspects. It identifies dynamic behavior of variable length without pre-training, operating in a continual learning manner. Real-time data preprocessing and bottom-up construction of a Behavior Forest provides an analytical representation of \ac{dvb}, enabling effective data recording based on data prevalence as a relevance criterion. Numerical examples across a synthetic and an ECG data set showcase the method's efficacy in discovering patterns. In a vehicle data set, 822 distinct \ac{dvb} were identified, reducing data by $\SI{96.01}{\percent}$,  while relevant \ac{dvb} remain included. Future research into selecting vehicle data based on multivariate time series using Behavior Forests and sharing them for faster trigger convergence shows promise for fleet coverage analysis, potentially integrating additional sensor modalities such as camera data analysis to enhance analysis capabilities.

\section{ACKNOWLEDGMENT}
This work results from the just better DATA (jbDATA)
project supported by the German Federal Ministry for Economic Affairs and Climate Action of Germany (BMWK) and the European Union, grant number 19A23003H.

%%%%%%%%%%%%%%%%%%%%%%%%%%%%%%%%%%%%%%%%%%%%%%%%%%%%%%%%%%%%%%%%%%%%%%%%%%%%%%%%

%References are important to the reader; therefore, each citation must be complete and correct. If at all possible, references should be commonly available publications.

\bibliographystyle{IEEEtran}
\bibliography{IEEEabrv,root}

\end{document}